\documentclass[sigconf]{acmart}

\usepackage{balance}

\def\BibTeX{{\rm B\kern-.05em{\sc i\kern-.025em b}\kern-.08emT\kern-.1667em\lower.7ex\hbox{E}\kern-.125emX}}
    
\copyrightyear{2019} 
\acmYear{2019} 
\setcopyright{acmlicensed}
\acmConference[CIKM '19]{The 28th ACM International Conference on Information and Knowledge Management}{November 3--7, 2019}{Beijing, China}
\acmBooktitle{The 28th ACM International Conference on Information and Knowledge Management (CIKM '19), November 3--7, 2019, Beijing, China}
\acmPrice{15.00}
\acmDOI{10.1145/3357384.3358099}
\acmISBN{978-1-4503-6976-3/19/11}

\usepackage{CJKutf8}
\usepackage{array}
\newcommand{\PreserveBackslash}[1]{\let\temp=\\#1\let\\=\temp}
\newcolumntype{C}[1]{>{\PreserveBackslash\centering}p{#1}}
\newcolumntype{R}[1]{>{\PreserveBackslash\raggedleft}p{#1}}
\newcolumntype{L}[1]{>{\PreserveBackslash\raggedright}p{#1}} 

\begin{document}

\fancyhead{}

\title[Query-bag Matching]{Query-bag Matching with Mutual Coverage for Information-seeking Conversations in E-commerce}

\author{Zhenxin Fu}
\authornote{This work was done when Zhenxin Fu was an intern at Alibaba Group.}
\email{fuzhenxin@pku.edu.cn}
\affiliation{%
  \institution{WICT, Peking University}
}

\author{Feng Ji}
\email{zhongxiu.jf@alibaba-inc.com }
\affiliation{%
  \institution{DAMO Academy, Alibaba Group}
}

\author{Wenpeng Hu}
\email{wenpeng.hu@pku.edu.cn}
\affiliation{%
  \institution{WICT, Peking University}
}

\author{Wei Zhou}
\email{fayi.zw@alibaba-inc.com }
\affiliation{%
  \institution{DAMO Academy, Alibaba Group}
}

\author{Dongyan Zhao}
\email{zhaodongyan@pku.edu.cn}
\affiliation{%
  \institution{WICT, Peking University}
}

\author{Haiqing Chen}
\email{haiqing.chenhq@alibaba-inc.com }
\affiliation{%
  \institution{DAMO Academy, Alibaba Group}
}

\author{Rui Yan}
\email{ruiyan@pku.edu.cn}
\authornote{Corresponding author: Rui Yan (ruiyan@pku.edu.cn)}
\affiliation{%
  \institution{WICT, Peking University}
}

%
\renewcommand{\shortauthors}{Fu et al.}

%
\begin{abstract}
Information-seeking conversation system aims at satisfying the information needs of users through conversations.
Text matching between a user query and a pre-collected question is an important part of the information-seeking conversation in E-commerce. In the practical scenario, a sort of questions always correspond to a same answer. Naturally, these questions can form a bag. Learning the matching between user query and bag directly may improve the conversation performance, denoted as query-bag matching. Inspired by such opinion, we propose a query-bag matching model which mainly utilizes the mutual coverage between query and bag and measures the degree of the content in the query mentioned by the bag, and vice verse. In addition, the learned bag representation in word level helps find the main points of a bag in a fine grade and promotes the query-bag matching performance. Experiments on two datasets show the effectiveness of our model.\footnotetext{WICT is the abbreviation for ``Wangxuan Institute of Computer Technology''.} \footnotetext{This work was supported by the National Key R\&D Program of China (No. 2017YFC0804001), the National Science Foundation of China (NSFC No. 61876196 and NSFC No. 61672058). Rui Yan was sponsored by Alibaba Innovative Research (AIR) Grant.}
\end{abstract}


\maketitle

\section{Introduction}
AliMe\footnote{\url{http://www.alixiaomi.com}} Bot is a kind of retrieval-based online service of E-commerce which collects a lot of predefined question-answering pairs. Through data analysis, we find that many variants of a question exist which means a sort of questions can correspond to a same answer. Based on the observation, naturally, we can view these questions with the same answer as a bag. Obviously, the bag contains diverse expressions of a question, which may provide more matching evidence than only one question due to the rich information contained in the bag. Motivated by the fact, different from existing query-question (Q-Q) matching method, we propose a new query-bag matching approach for retrieval-based chatbots. Concretely, when a user raises a query, the query-bag matching model provides the most suitable bag and returns the corresponding answer of the bag. To our knowledge, there is no query-bag matching study exists, and we focus on the new approach in this paper.

Recalling the text matching task \cite{yang-etal-2019-simple}, recently, researchers have adopted the deep neural network to model the matching relationship. ESIM \citep{chen2016enhanced} judges the inference relationship between two sentences by enhanced LSTM and interaction space. SMN \citep{wu2016sequential} performs the context-response matching for the open-domain dialog system. \citeauthor{W17-4121} \citep{W17-4121} explores the usefulness of noisy pre-training in the paraphrase identification task. \citeauthor{li2012machine} \citep{li2012machine} surveys the methods in query-document matching in web search which focuses on the topic model, the dependency model, etc. However, none of them pays attention to the query-bag matching which concentrates on the matching for a query and a bag containing multiple questions.

When a user poses a query to the bot, the bot searches the most similar bag and uses the corresponding answer to reply to the user. The more information in the query covered by the bag, the more likely the bag's corresponding answer answers the query. What's more, the bag should not have too much information exceeding the query. Thus modelling the bag-to-query and query-to-bag coverage is essential in this task.

In this paper, we propose a simple but effective mutual coverage component to model the above-mentioned problem. The coverage is based on the cross-attention matrix of the query-bag pair which indicates the matching degree of elements between the query and bag. The mutual coverage is performed by stacking the cross-attention matrix along two directions, i.e., query and bag, in the word level respectively. In addition to the mutual coverage, a bag representation in word level is issued to help discover the main points of a bag. The bag representation then provides new matching evidence to the query-bag matching model.

We conduct experiments on the AliMe and Quora dataset for the query-bag matching based information-seeking conversation. Compared with baselines, we verify the effectiveness of our model. Our model obtains 0.05 and 0.03 $\text{R}_{10}@1$ gains comparing to the strongest baseline in the two datasets. The ablation study shows the usefulness of the components. The contributions in this paper are summarized as follows: 1) To the best of our knowledge, we are the first to adopt query-bag matching in the information-seeking conversation. 2) We propose the mutual coverage model to measure the information coverage in the query-bag matching. 3) We release the composite Quora dataset to facilitate the research in this area.

\begin{figure}[!tp]
    \centering
    \includegraphics[width=1.0\columnwidth]{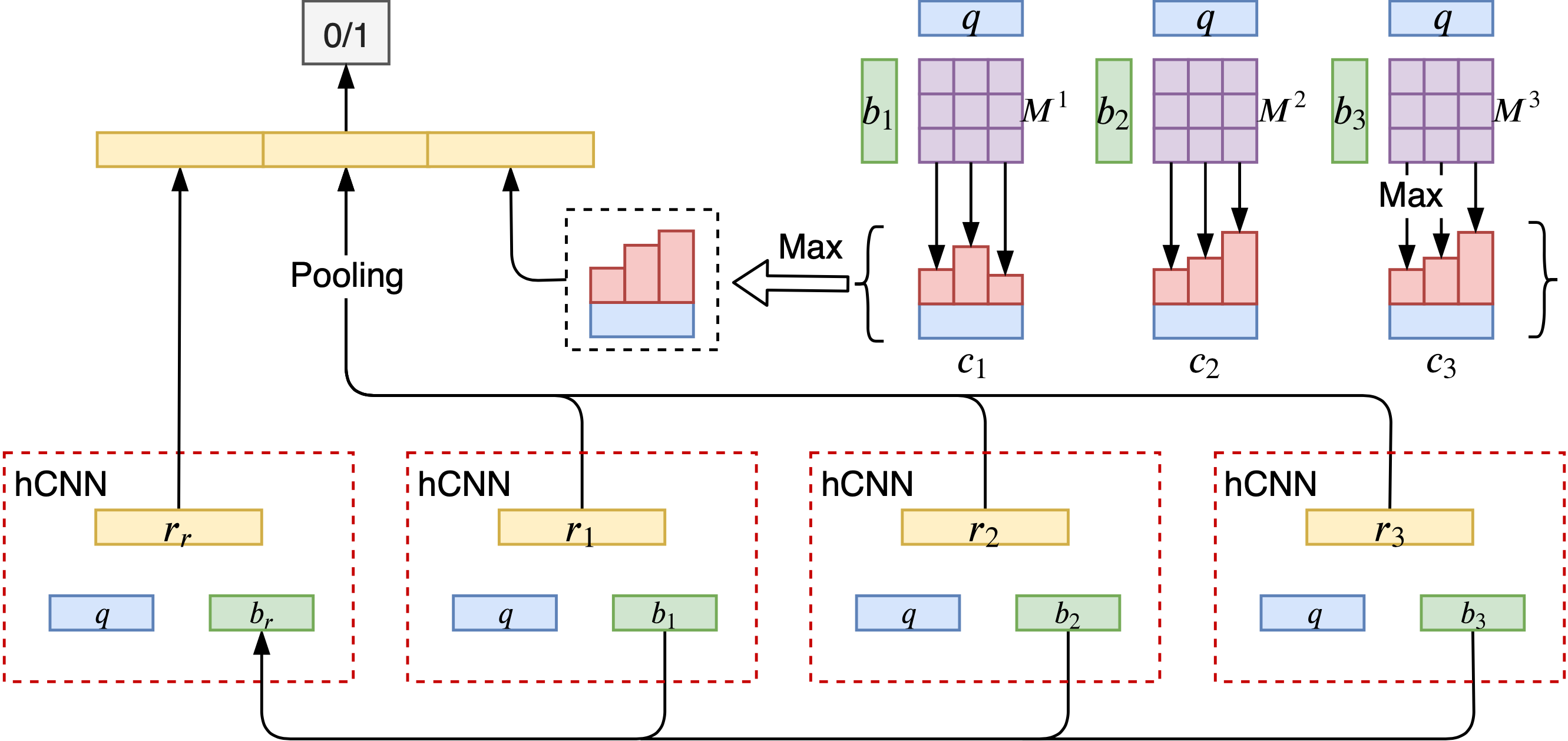}
    \caption{The proposed query-bag matching (QBM) model. The upper right is the query-bag coverage component (we only show the bag-to-query coverage in the Figure for demonstration, and the query-to-bag coverage is similar with bag-to-query coverage). q is the query, and $b_i$ is the $i$-th question in the bag. $M^i$ is the cross-attention matrix between $q$ and $b_i$. The bottom lines indicate the TF-IDF based bag representation construction. $b_r$ is a new ``question'' for bag representation.}
    \label{fig:model}
\end{figure}

\section{Methodology}
This task aims at predicting whether a query $q$ matches a bag $b$, where the bag is composed of some questions $b=\{ b_1, \dots, b_n \}$ and $n$ is the number of questions in a bag. For the $q$ and $b_i$, an embedding layer is first applied to transform words to word embeddings via looking up word embedding table which is initialized by pre-trained word embeddings as in Section \ref{sec:setup}.

In the following subsections, we will introduce our proposed Query-bag Matching (QBM) model which output is the matching probability indicating whether the query and bag are asking the same questions. The basic Q-Q (query-question) matching model hybrid CNN (hCNN) \citep{yu2018modelling} is presented as the background. Then we will show the base model and its two components designed to promote the performance: Mutual Coverage and Bag Representation. For better understanding, the whole model is shown in Figure \ref{fig:model}.

\subsection{Background: hCNN for Q-Q Matching}
We adopt the hCNN model, which measures the relationship between query-question pairs, to obtain the Q-Q matching representation. The model can be easily adapted to other query-question matching models.
hCNN is a CNN based matching model which is fast enough to work on the industry application. The input of hCNN is a query $q$ and the $i$-th question $b_i$ in the bag. $q$ and $b_i$ are fed into a CNN respectively. A cross-attention matrix $M^i$ is fed into another CNN to get the interaction representation between them. Each element of $M^i$ is defined as $M^i_{a,b}=q_a^\top \cdot b_{i,b}$ where $q_a$ is the word embedding of the $a$-th word in query $q$ and $b_{i,b}$ is the embedding of the $b$-th word in $b_i$. Finally, the outputs of CNNs are combined via Equation \ref{hcnn2} to get the representation $r_i$, which indicates the matching representation of the query $q$ and  the $i$-th question $b_i$ in the bag. For the Q-Q matching task, the $r_i$ is fed into an MLP (Multi-Layer Perceptron) to predict the matching score. In our query-bag matching setting, we will aggregate the $\{r_1, \dots, r_n\}$ to predict the query-bag matching score. Due to the page limitation, please refer to \citeauthor{yu2018modelling} \citep{yu2018modelling} for more details on hCNN.
\begin{align}
h_1 = \text{CNN}_1(q) \quad &  h_2^i = \text{CNN}_1(b_i) \quad h_m^i = \text{CNN}_2(q^\top \cdot b_i)  \\
r_i =& [h_1; h_2^i; h_1-h_2^i; h_1 \cdot h_2^i;h_m^i]    \label{hcnn2}
\end{align}

\subsection{Base Model}
After getting the Q-Q matching representation $r_i$, we combine the $\{r_1, \dots, r_n\}$ by element-wise max and mean pooling in order to get $r_p$ to represent the query-bag matching representation:
\begin{align}
    r_p = [ \text{max\_pooling} \{ r_1, \dots, r_n \}; \text{mean\_pooling} \{ r_1, \dots, r_n \} ] \label{rp}
\end{align}
where [;] denotes concatenation. After that, an MLP with softmax is applied to predict whether the query and the bag is asking the same question. Finally, the loss function minimizes the cross entropy of the training data.
Due to the out-of-order of the bag, we do not model the bag representation by CNN or LSTM, and experiments show the pooling-based method works well.

\subsection{Mutual Coverage}
\emph{``How many parts of a query are covered by the bag?''} and \emph{``Is all the information in the bag mentioned by the query?''} are two important problems in the query-bag matching task. We propose a novel mutual coverage module to model the above-mentioned inter-projection problems.

\textbf{Bag-to-query} Considering the $i$-th question $b_i$ in the bag, 
the element-wise max pooling is performed on $\{ M^i_0, \cdots M^i_n \}$ 
to get the $b_i$ to $q$ coverage $c_i=\text{max\_pooling}\{ M^i_0, \cdots M^i_n \}$ where 
$M^i$ is the cross-attention matrix between $b_i$ and $q$ as in the background section,
and $M^i_j$ is its $j$-th row. 
Each element $c_{i,j}$ represents how many information of the $j$-th word in $q$ is mentioned by the $i$-th question in the bag. To get the coverage from a bag instead of the $i$-th question in a bag, a new element-wise max pooling is applied on the generated $\{ c_1, \dots, c_n \}$ to get bag-to-query coverage $c_q$. The process of the bag-to-query coverage is shown in Figure \ref{fig:model}.

\textbf{Query-to-bag} The query-to-bag coverage is performed in a similar way. After getting the coverage $c_i$ from query $q$ to $b_i$. The concatenation of $\{ c_1, \dots, c_n \}$ across all questions in a bag forms the query-to-bag coverage vector $c_b$.

In addition, not all words in a question should be treated equally. The word ``the'' contributes little to the matching degree in most cases. However, ``package'' is very important in the E-commerce scenario. We adopt the attention mechanism \citep{bahdanau2014neural} to weight the coverage vector $c_q$ and $c_b$. The attention is calculated as follows (we take the bag-to-query coverage as an example):
\begin{align}
    e_j = \text{MLP}(q_j)  \qquad   \bar{e}_j  \propto \exp \{ e_j \}  \qquad  \bar{c}_q = \bar{e} \cdot c_q  \label{equ:att}
\end{align}
where $q_j$ is the embedding of $j$-th word in query. And the weighting of query-to-bag coverage performs in the same way. We call the mechanism coverage weighting.

The query-to-bag coverage, and bag-to-query coverage representation, and their summation are concatenated to the matching representation $r_p$ to predict the final score:
\begin{align}
    [ r_p ; \bar{c}_q ; \bar{c}_b ; \text{sum}(\bar{c}_q) ; \text{sum}(\bar{c}_b)]
\end{align}

\begin{table*}[]
    \centering
    \caption{
    Results of models and baselines with ablation study.
    MC and BR denote Mutual Coverage and Bag Representation respectively. ``BR w/o Cov'' denotes Bag Representation component without coverage module. \ddag ~and \S ~means the results are significant with $\text{p-value}<0.05$ measured by the Student’s paired t-test over the best baseline and the base model respectively. }
    \begin{tabular}{|l|lllll|lllll|}
    \hline
                   & \multicolumn{5}{c|}{AliMe}      &    \multicolumn{5}{c|}{Quora}     \\ \hline
        Model      & MRR &$\text{R}_{10}@1$ & $\text{R}_{10}@2$ & $\text{R}_{10}@5$ & $\text{R}_2@1$& MRR &$\text{R}_{10}@1$ & $\text{R}_{10}@2$ & $\text{R}_{10}@5$ & $\text{R}_2@1$  \\ \hline
        Q-Q Mean   & 0.6122 & 0.5050 & 0.5623 & 0.7287 & 0.8473 & 0.8350 & 0.7847 & 0.8133 & 0.8973 & 0.9480 \\
        Q-Q Max    & 0.6470 & 0.5477 & 0.6000 & 0.7590 & 0.8523 & 0.8438 & 0.7980 & 0.8227 & 0.8980 & 0.9493 \\ \hline   
        Bag-Con    & 0.6552 & 0.5610 & 0.6087 & 0.7607 & 0.8553 & 0.8026 & 0.7420 & 0.7800 & 0.8740 & 0.9287 \\ \hline
        Base       & 0.6845 & 0.6027 & 0.6397 & 0.7700 & 0.8707 & 0.8184 & 0.7643 & 0.7973 & 0.8800 & 0.9337 \\ 
        Base+MC   & 0.6936 & 0.6137 & 0.6497 & 0.7807 & \textbf{0.8823} & 0.8640 & 0.8247 & 0.8480 & 0.9083 & \textbf{0.9587} \\ 
        Base+BR   & 0.6913 & 0.6103 & 0.6443 & 0.7833 & 0.8783 & 0.8628 & 0.8213 & 0.8477 & 0.9123 & 0.9497 \\ 
        Base+(BR w/o Cov) & 0.6849 & 0.6013 & 0.6410 & 0.7810 & 0.8727 & 0.8280 & 0.7763 & 0.8093 & 0.8833 & 0.9430 \\
        QBM (Base+BR+MC) & \textbf{0.7007}$^{\ddag\S}$ & \textbf{0.6197}$^{\ddag\S}$ & 0.\textbf{6600}$^{\ddag\S}$ & \textbf{0.7923}$^{\ddag\S}$ & \textbf{0.8823}$^{\ddag}$ & \textbf{0.8656}$^{\ddag\S}$ & \textbf{0.8253}$^{\ddag\S}$ & \textbf{0.8510}$^{\ddag\S}$ & \textbf{0.9137}$^{\S}$ & 0.9520$^{\ddag\S}$ \\  \hline
    \end{tabular}
    \label{tab:res}
\end{table*}

\subsection{Bag Representation}
All the questions in a bag follow the same question points because they are different variants of the same question. We model the question points by collecting the important words in the bag,  forming the word-level bag representation.
We collect the top-10 important words through TF-IDF algorithm, except stop words, in a bag to form a new ``question'' $b_r$, and an hCNN is used to model the relationship of the user query and the new ``question'' $b_r$ in order to obtain the matching representation $r_r$. The $r_r$ is then concatenated to the matching representation $r_p$ as a new feature to predict the query-bag matching degree. We also adopt the coverage mechanism discussed above over the cross-attention matrix between the query and the new ``question''. The new coverage representation is also concatenated to the $r_p$.

\section{Experiments}
\subsection{Dataset}
We conduct experiments on two datasets: AliMe and Quora. The AliMe dataset is collected from the AliMe intelligent assistant system and the Quora dataset is composed of a public dataset.

\textbf{AliMe}
For the AliMe service in E-commerce, we collect 8,004 query-bag pairs to form our dataset. Negative sampling is also an important part of the matching model. For each query, we use the Lucene\footnote{\url{http://lucene.apache.org}} to retrieval the top-20 most similar questions from the whole question candidates. Then we filter the questions which are in the corresponding right bag. After that, we randomly sample one in the retrieved candidate and use the bag that the retrieved candidate belongs to as the negative case. In the bag construction stage, the annotators have already merged all the questions of the same meaning, so we can ensure that the after filtering retrieved cases are negative in our setting. We also restrict the number of questions in a bag not more than 5 and discard the redundant questions. Finally, we get 12,008 training cases, 2,000 valid cases, and 10,000 test cases. Please notice, for the testing, we sampled 9 negative bags instead of 1, and thus formed 10 candidates for ranking.

\textbf{Quora}
The Quora dataset is originally released for the duplicated question detection task\footnote{\url{https://data.quora.com/First-Quora-Dataset-Release-Question-Pairs}}. The dataset contains 400,000 question pairs and each pair is marked whether they are asking the same question. Due to the huge amount of duplicated question pairs, we group the questions as question bag via the union-find algorithm from the duplicated questions. We get 60,400 bags, and all the questions in a bag are asking the same question. We filter the bags that contain questions less than 3 to make the bag not too small. The new bag dataset will help similar questions recommendation on the Quora website. We then extract one question in the bag as query and the other questions make up the bag in our task. Considering the negative samples, we follow the same strategy as AliMe dataset. Finally, we get 20,354 training set, 2,000 validation set, and 10,000 test set. To facilitate the research in this area, the  composed Quora dataset are released\footnote{The composed Quora dataset and more detailed experiments are in \url{https://github.com/fuzhenxin/Query-Bag-Matching-CIKM}}.

\subsection{Setup}
\label{sec:setup}
We use the Adam optimizer with learning rate 0.0001 to optimize the parameters. The batch size is 32. The dropout rate is 0.5. The max length of the query and questions is 20 to cover most of the words in a sentence. We use padding to handle the various lengths of the text. The model checkpoint is chosen according to the best F-score on the validation set. The word embedding dimension is 300, and the pre-trained word embedding is from Sina\footnote{\url{https://github.com/Embedding/Chinese-Word-Vectors}} 
and Glove\footnote{\url{https://nlp.stanford.edu/projects/glove}} for AliMe and Quora dataset respectively. Besides, the embedding is tuned while the model training to get better performance.

\subsection{Baselines}
To prove the effectiveness of our models, we propose two baselines from different aspects: the Q-Q matching based baseline and the query-bag matching based baseline.

\textbf{Q-Q Matching}
One starting point behind our work is that the query-bag matching may work better than the Q-Q matching for the information-seeking conversation. To verify such opinion, we propose the Q-Q matching based baseline and compare our model with two instances of the baseline. 
We extract the query-question pairs form the query-bag pair. The label of the query-bag pair is assigned to the new query-question pairs.
An hCNN model is applied to train the new dataset. In the testing stage, each query-question pair is assigned with a probability indicating the matching degree. To compare with our model, we rank the bags based on the query-bag matching scores and the scores are defined as the max or mean matching probability of the query-question pairs in the query-bag pair. We name the two instances Q-Q Max and Q-Q Mean respectively.

\textbf{Query-bag Matching}
To verify the effectiveness of our proposed models, We design a new query-bag matching based baseline. We concatenate the questions in the bag to form a new long ``question'', then the hCNN model is applied to measure the matching degree of the original query and the new ``question'', namely Bag-Con (Bag Concatenation). 

\subsection{Evaluation}
Following \citet{qiu2018transfer}, we evaluate the model performance on five automatic evaluation metrics: MRR, $\text{R}_{10}@1$, $\text{R}_{10}@2$, $\text{R}_{10}@5$, and $\text{R}_{2}@1$. $\text{R}_n@k$ calculates the recall of the true positive pre-defined questions among the $k$ selected candidates from $n$ available candidates. And Mean Reciprocal Rank (MRR) is another popular measurement for ranking problems.

\section{Results and Analysis}
\textbf{Results and Ablation Study} The results are shown in Table \ref{tab:res}. Our model (QBM) performs best compared to baselines (Q-Q Mean, Q-Q Max, Bag-con). Comparing Bag-Con and Base model, we find that modelling the query-question relationship following aggregation works better. We assume that the pooling-based aggregation can reduce the redundant information cross sentences in a bag. Considering the Q-Q matching based methods and query-bag based methods. In AliMe dataset, the query-bag matching outperforms the Q-Q matching based methods which shows the necessity to perform query-bag matching.
The ablation study shows that the mutual coverage component and bag representation component achieve better performance than the base model, especially in the Quora dataset. The two components work independently and their combination gets the best performance. 

\begin{CJK}{UTF8}{gbsn}
\begin{table}[]
    \centering
    \caption{Some words and their corresponding weights ($e$ in Equation \ref{equ:att}) in mutual coverage module. The average weight across the whole vocabulary is also presented here.}
    \begin{tabular}{|L{2.5cm}|L{1cm}|L{1.5cm}|L{1cm}|}
    \hline
        \multicolumn{2}{|c|}{AliMe}  &   \multicolumn{2}{c|}{Quora} \\ \hline
        的 ('s)           & 1.180    & The      & 0.006                \\ 
        和 (And)          & 1.237    & And      & 0.894                \\ 
        退款 (Refund)     & 5.042    & Where    & 1.366                \\ 
        机票 (Air ticket) & 6.484    & America  & 2.018                \\ \hline
        Average           & 2.202    & Average  & 0.899                \\ \hline
    \end{tabular}
    \label{tab:coverage_weight}
\end{table}
\end{CJK}

\textbf{Effectiveness of the Mutual Coverage} To intuitively learn the coverage weight, we sample some words with their weights in Table \ref{tab:coverage_weight}. It shows that the words like ``The'' have low weight, which confirms that they contribute little to the matching. ``Refund'' in E-commerce is a very important element in a user query sentence. And ``America'' is important in Quora, because question like ``what is the capital in <location>?'' is highly related to location ``<location>''.

\textbf{Analysis of the Bag Representation} Coverage is also applied in the bag representation layer. The results of the bag representation without coverage component (Base+(BR w/o Cov)) is shown in Table \ref{tab:res}. Compared with the Base+BR and BR without coverage, it shows that the coverage component contributes a lot on both the two datasets. The bag representation with coverage (Base+BR) gains improvement over Base model, especially in Quora dataset.

\section{Conclusion}
In this paper, we propose the QBM model which performs the query-bag matching in information-seeking conversation. Experiments show that the proposed mutual coverage component improves the model performance. And the model can automatically discover important words in the query or bag from both the coverage weighting component and the word-level bag representation.
This work also shows that learning the query-bag matching directly in some scenarios may outperform the query-question matching in ranking bags. 
One advantage of our model is that it is extensible in replacing the query-question matching component.

\bibliographystyle{ACM-Reference-Format}
\bibliography{sample-base}

\end{document}